\pdfoutput=1

\documentclass[11pt]{article}

\usepackage{ACL2023}

\usepackage{times}
\usepackage{latexsym}

\usepackage[T1]{fontenc}

\usepackage[utf8]{inputenc}

\usepackage{microtype}

\usepackage{inconsolata}

\usepackage{graphicx}
\usepackage{booktabs}
\usepackage{tikz}
\usepackage{pgfplots} 
\usepackage{scalefnt} 
\usepackage{caption}
\usepackage{subcaption}
\usepackage{enumitem}
\usepackage{todonotes}

\title{Balanced Data Sampling for Language Model Training with Clustering}

\author{
    Yunfan Shao$^{1,2*}$,
    Linyang Li$^{1,2}$\thanks{\ \ Equal Contribution},
    Zhaoye Fei$^{1,2}$,
    Hang Yan$^{2,3\dagger}$,
    Dahua Lin$^{2,3}$,
    Xipeng Qiu$^{1}$\thanks{\ \ Corresponding Author.} \\
    $^1$School of Computer Science, Fudan University\\
    $^2$Shanghai AI Laboratory\\
    $^3$The Chinese University of Hong Kong \\
    \texttt{\{yfshao19, zyfei20, xpqiu\}@fudan.edu.cn} \\
    \texttt{\{lilinyang, yanhang\}@pjlab.org.cn}
}

\begin{document}
\maketitle
\begin{abstract}

Data plays a fundamental role in the training of Large Language Models (LLMs). While attention has been paid to the collection and composition of datasets, determining the data sampling strategy in training remains an open question. Most LLMs are trained with a simple strategy, random sampling. However, this sampling strategy ignores the unbalanced nature of training data distribution, which can be sub-optimal. In this paper, we propose ClusterClip Sampling to balance the text distribution of training data for better model training. 
Specifically, ClusterClip Sampling utilizes data clustering to reflect the data distribution of the training set and balances the common samples and rare samples during training based on the cluster results. A repetition clip operation is introduced to mitigate the overfitting issue led by samples from certain clusters.
Extensive experiments validate the effectiveness of ClusterClip Sampling, which outperforms random sampling and other cluster-based sampling variants under various training datasets and large language models~\footnote{Our code is released at \url{https://github.com/choosewhatulike/cluster-clip}}. 

\end{abstract}

\section{Introduction}

Large Language Models (LLMs) have opened new frontiers in understanding and generating human languages~\cite{brown2020gpt3,touvron2023llama,openai2023gpt4}.
A critical aspect of training these models lies in the acquisition and organization of training data~\cite{wang2023dataSurvey}.
Some works focus on selecting high-quality data. These works usually perform data filtering or data cleaning from a large corpus, based on either rule-based~\cite{gao2021pile,together2023redpajama} or model-based algorithms~\cite{abbas2023demdedup,tirumala2023d4,marion2023prunePretrain}. On the other hand, several approaches concentrate on optimizing the composition weights of the collected data. These methods typically focus on optimizing the domain weights either using heuristics~\cite{du2022galm,touvron2023llama} or model statistics~\cite{xie2023doremi,fan2023doge}. 
While much attention has been given to the collection and composition of diverse datasets for training LLMs~\cite{gao2021pile,together2023redpajama,touvron2023llama2,mukherjee2023orca}, it is still unclear how the training data sampling affects the optimization of language models. 

The sampling methods of existing works are coarse-grained, which determines the sampling weights or sampling order of each domain. The \textit{domain} is usually defined based on the data source or other metadata during the dataset collection, which is coarse-grained and inaccurate. For instance, the Llama models~\cite{touvron2023llama} assign domain mixture weights heuristically, including 67\% CommonCrawl web data, 4.5\% Github code, 4.5\% Wikipedia documents, etc. And ~\citet{roziere2023codeLlama,azerbayev2023llemma} improve certain abilities of LLMs by tuning the model on specific domains, like code or mathematical texts.
However, the texts are sampled randomly in each domain, which ignores the unbalanced distribution of the expressed meanings and topics. Due to the nature of the data corpus, texts with similar meanings have a long tail distribution in the training set~\cite{Zipf1949HumanBA,Chan2022distributionICL,abbas2023demdedup}. When using random sampling, LLMs can underfit rare documents and overfit common samples. It is straightforward to utilize a uniform sampling strategy that samples texts with different meanings evenly, which up-samples rare documents and down-samples common texts. However, uniform sampling will up-weight rare documents and repeat them so many times in the training, resulting in severe overfitting of trained LLMs on these training samples.
\begin{figure*}[htbp]
    \centering
    \includegraphics[width=0.8\textwidth]{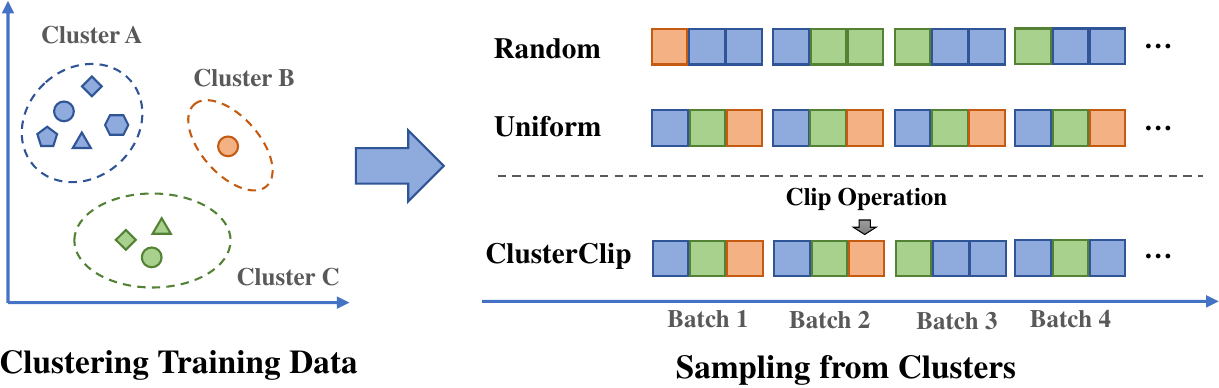}
    \caption{Illustration of \textbf{ClusterClip Sampling}. The algorithm utilizes data clustering to describe the training data distribution. Then, it balances the sampling probabilities of samples in different clusters during the training. Moreover, a clip operation is introduced to knock out samples with too many repetitions.}
    \label{fig:multi-sampling-strategy}
\end{figure*}

To address the above challenges, we propose \textbf{ClusterClip}, a cluster-based sampling strategy with clip operation to mitigate overfitting. This sampling strategy has two steps. Firstly, we leverage data clustering to reflect the data distribution. Based on off-the-shelf NLP tools, the semantic-related texts could be grouped into the same cluster. By calculating the size of each semantic cluster, we can evaluate the data rarity. Secondly, we perform the data sampling using the cluster information. The documents from different clusters are sampled evenly at the beginning of the training, thus encouraging the model to learn from rare documents instead of wasting computation on common texts. As the training progresses, a clip operation is applied. If certain documents are sampled too many times, these documents are clipped and no longer sampled from the dataset. Thus, ClusterClip rebalances the data distribution, which facilitates the model to learn rare documents but avoids severe overfitting of repeated texts by clipping.

Our approach is distinguished by its ability to improve model learning efficiency and generalization without relying on dataset-specific metadata or complicated optimizations. 
Extensive experiments demonstrate the versatility and effectiveness of our proposed ClusterClip Sampling. We show that it consistently enhances the performance of representative LLMs, Llama2-7B~\cite{touvron2023llama2} and Mistral-7B~\cite{jiang2023mistral7b}, in both pre-training and supervised fine-tuning scenarios, indicating its broad applicability. Moreover, we further evaluate several variants of cluster-based sampling methods and demonstrate that: (1) All the cluster-based sampling method variants outperform random sampling, indicating the effectiveness of sampling based on semantic distribution;
(2) The Clip operation effectively mitigates the overfitting of repeated documents, leading to large improvements in diverse tasks.
(3) Specializing data training order (e.g. General-to-Specific or Specific-to-General) can affect the performance of LLMs on various downstream tasks, showing the promise of progressive learning on LLMs; 

To sum up, our contributions are as follows:
\begin{itemize}
    \item We propose ClusterClip Sampling, which rebalances the occurrence of common or rare documents with a clip operation to avoid severe overfitting. 
    \item We validate the effectiveness of ClusterClip Sampling on multiple datasets and LLMs, demonstrating the broad applicability and stable improvements across pre-training and fine-tuning. 
    \item We present representative variant cluster-based sampling methods to show where the improvements of ClusterClip come from and provide interesting results for future design of sampling strategies.
\end{itemize}

\section{Related Works}
\paragraph{Data Clustering for LLMs}
Clustering methods are widely applied in data curation of LLM pre-training. A line of work incorporates clustering for data deduplication. Semdedup~\cite{abbas2023demdedup} aims to eliminate semantic duplicates from the pre-training corpus and use clustering in the sentence embedding space to reduce the computational cost. D4~\cite{tirumala2023d4} further aggressively removes duplicate documents by combining multiple clustering-based deduplication methods.
Moreover, clustering also has been used to improve data quality. Existing works perform quality filtering using a classifier to find data samples that are close to high-quality texts~\cite{brown2020gpt3,gao2021pile,du2022galm,touvron2023llama},. And MiniPile~\cite{kaddour2023minipile} is constructed by filtering out low-quality 
clusters based on semantic embeddings. 
Different from previous works, we explore the effectiveness of clustering for learning strategies instead of data curation for LLM training.

\paragraph{Data Composition for LLM Pre-Training}
The composition of pre-training data is a determinant of LLM performance. 
Efforts are made to collect and utilize domain mixtures of pre-training data to improve LLM performance~\cite{longpre2023pretrainGuide,shen2023slimPajama,nijkamp2023codegen2}. \citet{brown2020gpt3,chung2022flanPalm,du2022galm,azerbayev2023llemma,touvron2023llama,touvron2023llama2} exploit manually designed domain composition weights by small-scale pre-training experiments.
However, we focus on cluster-based data sampling approaches, which do not rely on manual selection of data composition.
In addition, existing works also explore the algorithms for searching optimal domain weights. Several methods involve training auxiliary models, either proxy models or reference models to determine the composition weights~\cite{mindermann2022rhoLoss,xie2023doremi,fan2023doge,Suzuki2023extract}. Furthermore, some works concentrate on sample-level data selection by introducing model-based metrics to measure sample weights, which include importance score~\cite{xie2023dsir}, perplexity~\cite{xia2023shearedLlama}, and gradients~\cite{marion2023prunePretrain}. 
Different from these works, the cluster-based sampling strategies utilize off-the-shelf embedding models and semantic-based cluster-level data sampling.

\section{Methodology}
We propose \textbf{ClusterClip} Sampling, a cluster-based sampling strategy to rebalance the data distribution of the training corpus to facilitate model learning and mitigate overfitting. In this section, we delve into the detailed process of data clustering and the sampling strategy using cluster information to balance the sampling probabilities of common and rare documents.

\subsection{Data Clustering}
Aiming to describe and manipulate the distribution of texts in the training set, we introduce data clustering to group samples into semantic clusters. We choose not to rely on metadata from the dataset itself, as such information is often absent or fuzzy~\cite{gao2021pile, azerbayev2023llemma}. Instead, data clustering can automatically discover semantic similar documents and group these data points into clusters. Specifically, we first utilize off-the-shelf transformer-based models to generate text representations for each data sample. Then we conduct a K-Means clustering on these generated data embeddings to group samples into clusters. By clustering, we classify data with similar topics into the same subset. Thus, we can analyze the data distribution and rebalance the data distribution when sampling the training data. 

We choose out-of-the-box transformer-based embedding models and the K-Means method in the experiments as these methods are well-established, efficient at scale, and can produce semantic-related data clusters. Other embedding functions, including rule-based or model-based, could also be utilized for more accurate clustering. Comparing the impact of different embedding or clustering methods on the data sampling strategies would be a valuable topic and is our future work.

\subsection{ClusterClip Sampling}
After data clustering, the clusters can describe the long tail distribution of the training set. Based on the cluster information, ClusterClip Sampling increases the sampling weights of rare documents and decreases the weights of common texts. Moreover, a clip operation is introduced to mitigate overfitting. Thus, ClusterClip Sampling balances the learning on both very common texts and extremely rare documents.

\paragraph{Uniform Sampling}
At the beginning of training, ClusterClip Sampling performs a Uniform Sampling from the clusters, which aims to up-sample rare data points and down-sample common texts. We ensure that each cluster has the same probability of being sampled. After sampling the cluster, amount of tokens in each cluster. This also improves the data diversity within the batch as it balances the occurrence of samples in each cluster in a batch.

\paragraph{Clip Operation}
When uniformly sampling the data, documents from small clusters can be sampled a huge number of times. In this case, the model will suffer from overfitting on these small semantic clusters and not learn well on the whole training set. To solve this issue, we further propose a clip operation to add a maximum repetition of each sample. When different clusters are uniformly sampled, small clusters can be consumed multiple times. The ClusterClip will record the repeated times of each cluster. When one cluster has been consumed a certain number of times, the cluster will be knocked out and will not be sampled in further training. Thus, the model will see a sample at most a certain number of times, which mitigates the overfitting.

\section{Experimental Setups}
To validate the proposed ClusterClip Sampling, we conduct extensive experiments on multiple datasets and LLMs.
In this section, we introduce the experimental setups of our experiments, including the baselines, training datasets, training hyperparameters, and evaluation setups.

\subsection{Baselines}
To demonstrate the effectiveness of ClusterClip Sampling, we introduce several representative sampling methods for comparison. 
\paragraph{Random} The texts are sampled randomly, which is widely used in the pre-training and fine-tuning of many LLMs~\cite{touvron2023llama,touvron2023llama2,azerbayev2023llemma,mukherjee2023orca}.
\paragraph{Uniform} The texts are uniformly sampled from each cluster, which is a simplified version of ClusterClip without the clip operation.
\paragraph{General-to-Specific (G2S)} Inspired by recent practice of training domain-specific language models~\cite{roziere2023codeLlama, azerbayev2023llemma}, we want the model to learn general abilities before acquiring specific domain knowledge and skills. Thus, G2S is initiated by uniform sampling from each cluster. If a particular cluster is exhausted, sampling from that cluster ceases until the entire dataset has been traversed. 
By doing so, the model learns diverse and general samples before concentrating on some specific clusters. 
\paragraph{Specific-to-General (S2G)} The objective of S2G is to prioritize the model's learning of rare samples and can be viewed as the opposite of G2S. It involves initially training the model to acquire knowledge in some specific domains before learning general capabilities. To achieve this, it employs the exactly reversed sampling order of the G2S strategy for training. This can be related to progressive training or curriculum learning~\cite{bengio2009curriculum,hacohen2019curriculumDeep}, in which the model first learns from easy samples and then transfers to hard ones. 

\subsection{Training Datasets}
To fully validate the effectiveness of ClusterClip Sampling,
we train the models with different sampling methods, including the proposed method and baselines, on both supervised fine-tuning and pre-training setups. 

\paragraph{Superivsed Fine-Tuning Dataset}
We choose \textbf{Open-Orca}~\cite{lian2023openOrca} to probe the sampling strategies on supervised fine-tuning. It is an open-source implementation of Orca~\cite{mukherjee2023orca} that employs GPT-3.5 and GPT-4 to generate detailed answers and intermediate thoughts given a diverse set of NLP tasks. Following the methodology of Orca, Open-Orca collects 1 million outputs from GPT-4 and 3.2 million outputs from GPT-3.5 based on these inputs. The total size of the training set is about 1B tokens.

\paragraph{Pre-Training Dataset}
We utilized the \textbf{Proof-Pile-2} dataset~\cite{azerbayev2023llemma} to investigate the influence of various sampling strategies for continual pre-training on specific domains. The Proof-Pile-2 dataset is motivated by enhancing the mathematical reasoning capabilities of models and consists of three components: code files, web pages, and scientific papers, leading to 55B tokens in total.

\paragraph{Preprocessing and Clustering}
We use the base-sized Jina Embeddings 2\footnote{\url{https://huggingface.co/jinaai/jina-embeddings-v2-base-en}}~\cite{gunther2023jina2} with mean pooling to generate text embeddings. 
For Open-Orca, we do not differentiate between data generated by GPT-4 and GPT-3.5 and mix them for clustering and training, which reflects the results in a more realistic scenario with mixed instruction data quality. We concatenate the inputs and outputs and truncate the result text into 1024 tokens for embedding computation. Then, we run K-Means with cosine distance on generated embeddings over 300 iterations to obtain 2000 clusters. 
For Proof-Pile-2, we combined and shuffled different sub-domains of the mixture together and sampled 10B tokens as our training set. This approach allows us to explore the model behavior when changing data sampling strategies in the context of continual training on data mixtures with multiple domains.
After obtaining the document embeddings, we set the number of iterations and clusters of K-Means to 300 and 100 to obtain the clusters, respectively. 

\subsection{Training Setups}
We adopt representative LLMs for training using the InternLM~\cite{2023internlm}. All models are trained on 64 A800 GPUs with bfloat16 mixed precision.
\paragraph{Supervised Fine-Tuning Setups}
We choose Mistral-7B~\cite{jiang2023mistral7b} without aligning for chat as the backbone for supervised-fine-tuning. 
We fine-tuned the model with these sampling strategies separately, each for 20000 steps with a global batch size of 0.25 million tokens on Open-Orca, totaling 5B tokens. The context length is 4096 with packing. The learning rate is warmed up to $3e-6$ over the first 200 steps and then cosine decayed to $3e-7$ at the end of training. Following~\citet{Muennighoff2023dataRepeat}, the threshold of clipping is set to 5 for all experiments unless specified. To train one model, it utilizes 384 GPU hours.

\paragraph{Pre-Training Setups}
We initialize the backbone model from the Llama2-7B~\cite{touvron2023llama2} base model for continual pre-training.
We trained the model with each sampling method for 5000 steps with a global batch size of 4 million tokens, which used 20B tokens in total. The context length is 4096 with packing, and the learning rate is first warmed up to $5e-5$ over 200 steps and then cosine decayed to $1e-5$ at the end of training. The training utilizes 1216 GPU hours for each model.

\subsection{Cost Comparison}
We calculate the costs of clustering and training to figure out whether the clustering operation is a good investment when training a language model at scale. As shown in Table~\ref{tab:cluster-cost}, the cost of clustering is relatively low compared with the large language model (LLM) training, because the model to generate document embedding is very small (100M in our experiments), and the K-means algorithm is very fast by paralleling on GPUs. The Proof-Pile-2 is much faster because it contains much longer documents, e.g. Arxiv papers and web pages, and the clustering only considers the first 1024 tokens. Thus, the clustering cost mainly depends on the number of samples in datasets, instead of the actual tokens.

Moreover, clustering is a one-time investment, and we may train many models of varying scales once the training data is clustered, which can further amortize the clustering costs. To further speed up the clustering and lower the memory usage, one can use more efficient methods and shrink the embedding size as mentioned in ~\citet{Yamada2021retrieval,Kusupati2022Matryoshka}.

\begin{table*}[htbp]
    \centering
    \small
    \begin{tabular}{l|cccc}
        \toprule
        Dataset & Tokens & Samples & Avg. Length & Embedding Storage \\
        \midrule
        Open-Orca & 1.8B & 4.2M & 429 tokens & 14GB \\
        Proof-Pile-2 & 13.4B & 2.7M & 4963 tokens & 9GB \\
        \midrule
        Dataset & Embedding Cost & Kmeans Cost & Total Clustering Cost & Training 7B Cost \\
        \midrule
        Open-Orca & 6 GPU hours & 0.3 GPU hour & 6.3 GPU hours & 384 GPU hours \\
        Proof-Pile-2 & 5.5 GPU hours & 0.3 GPU hour & 5.8 GPU hours & 1216 GPU hours \\
        \bottomrule
    \end{tabular}
    \caption{Clustering and Training Costs of Datasets.}
    \label{tab:cluster-cost}
\end{table*}

\subsection{Evaluation Setups}
We introduce a diverse set of evaluation tasks and datasets to reflect the general and detailed model performance when incorporating different sampling methods.
\paragraph{Evaluation Datasets}
The trained models are evaluated on a wide range of downstream tasks, including SuperGLUE~\cite{wang2019superglue}, GSM8K~\cite{cobbe2012gsm8k}, MATH~\cite{hendrycks2021math}, OpenBookQA~\cite{mihaylov2018openbookqa}, MMLU~\cite{hendrycks2021mmlu}, BBH~\cite{suzgun2023bbh} and MT-Bench~\cite{zheng2023mtbench}.

\paragraph{Evaluation Methods}
Following~\citet{2023internlm}, we use perplexity-based evaluation for SuperGLUE, OpenBookQA, and MMLU. The few-shot chain-of-thought prompting~\cite{wei2022chainOfThought} is used to evaluate the accuracy of reasoning tasks, including GSM8K, MATH and BBH. The MT-Bench is evaluated by GPT-4 with reference-based scoring prompts~\cite{zheng2023mtbench}.

\section{Experimental Results}
In this section, we present the experimental results, including the comparison of the proposed sampling method with baselines (Sec~\ref{sec:main-results}), a detailed analysis of multiple variants of cluster-based sampling methods (Sec~\ref{sec:analysis-variants}), and the ablation study of different parts of ClusterClip (Sec~\ref{sec:ablation}).
\subsection{Main Results}\label{sec:main-results}
\paragraph{Supervised Fine-Tuning}
Experimental results on Open-Orca show the effectiveness of ClusterClip Sampling on the overall model performance across multiple domains and capabilities. 
As shown in Table~\ref{tab:compare-result-open-orca}, the model trained on Open-Orca with ClusterClip Sampling outperforms Uniform Sampling on SuperGLUE, OpenBookQA, and MT-Bench, and beats the Random Baselines with a large margin. The ClusterClip also achieves comparable performance when compared with the sampling methods G2S and S2G on Open-Orca. It demonstrates that the ClusterClip alleviates the overfitting issue of Uniform Sampling and improves the overall performance.
\begin{table}[htbp]
  \centering
    \resizebox{0.48\textwidth}{!}{
    \begin{tabular}{l|cccc}
    \toprule
          & \textbf{SuperGLUE} & \textbf{GSM8K} & \textbf{OpenBookQA} & \textbf{MT-Bench} \\
    \midrule
    \textbf{Mistral-7b} & 50.19  & 47.61  & 64.20  & - \\
    \midrule
    \textbf{Random} & 62.11  & 61.49  & 79.80  & 6.60  \\
    \textbf{Uniform} & 63.00  & 58.83  & 78.20  & 6.75  \\
    \textbf{G2S} & \textbf{65.41} & 59.36  & 79.40  & 6.81  \\
    \textbf{S2G} & 64.95  & \textbf{62.55} & 80.20 & \textbf{7.08} \\
    \midrule
    \textbf{ClusterClip} & 64.30 & 58.68 & \textbf{81.40} & 6.90 \\
    \bottomrule
    \end{tabular}%
    }
  \caption{Comparison of different sampling strategies on Open-Orca.}
  \label{tab:compare-result-open-orca}%
\end{table}%

\paragraph{Continual Pre-Training}

We also demonstrate the effectiveness of ClusterClip Sampling in continual pre-training. As shown in Table~\ref{tab:compare-result-proof-pile}, the model trained on Proof-Pile-2 with ClusterClip obtains strong overall performance, which achieves 7.90 on MATH and 51.05 on MMLU. This demonstrates that the ClusterClip Sampling outperforms other sampling methods on Proof-Pile-2. We also notice that the ClusterClip Sampling consistently outperforms Uniform Sampling both on Proof-Pile-2 and Open-Orca, which demonstrates that the overfitting issue is significant for Uniform Sampling and the Cutoff strategy alleviates this issue while still maintaining the benefits of data diversity of Uniform Sampling. 
Moreover, the G2S and S2G sampling methods underperform the ClusterClip method on all these downstream tasks, which are not consistent compared with results from supervised fine-tuning. It indicates that the effect of changing the training order is unstable. And ClusterClip Sampling is effective in both continual pre-training and supervised fine-tuning, demonstrating the generalization and board application.

\begin{table}[htbp]
  \centering
  \resizebox{0.42\textwidth}{!}{
    \begin{tabular}{l|cccc}
    \toprule
          & \textbf{MATH} & \textbf{GSM8K} & \textbf{MMLU} & \textbf{BBH} \\
          \midrule
    \textbf{LLama2-7B} & 3.50  & 16.68  & 46.79  & 38.20  \\
    \midrule
    \textbf{Random} & 6.52  & 25.55  & 48.84  & 41.81  \\
    \textbf{Uniform} & 7.62  & \textbf{26.00}  & 49.98  & \textbf{42.89}  \\
    \textbf{G2S} & 6.92  & 23.43  & 49.42  & 41.69  \\
    \textbf{S2G} & 6.98  & 23.12  & 48.61  & 40.90  \\
    \midrule
    \textbf{ClusterClip} & \textbf{7.90}  & 24.79  & \textbf{51.05}  & 42.78  \\
    \bottomrule
    \end{tabular}%
    }
  \caption{Comparison of different sampling strategies on Proof-Pile-2.}
  \label{tab:compare-result-proof-pile}%
\end{table}%

\begin{figure}[htbp]
    \centering
    \includegraphics[width=0.9\linewidth]{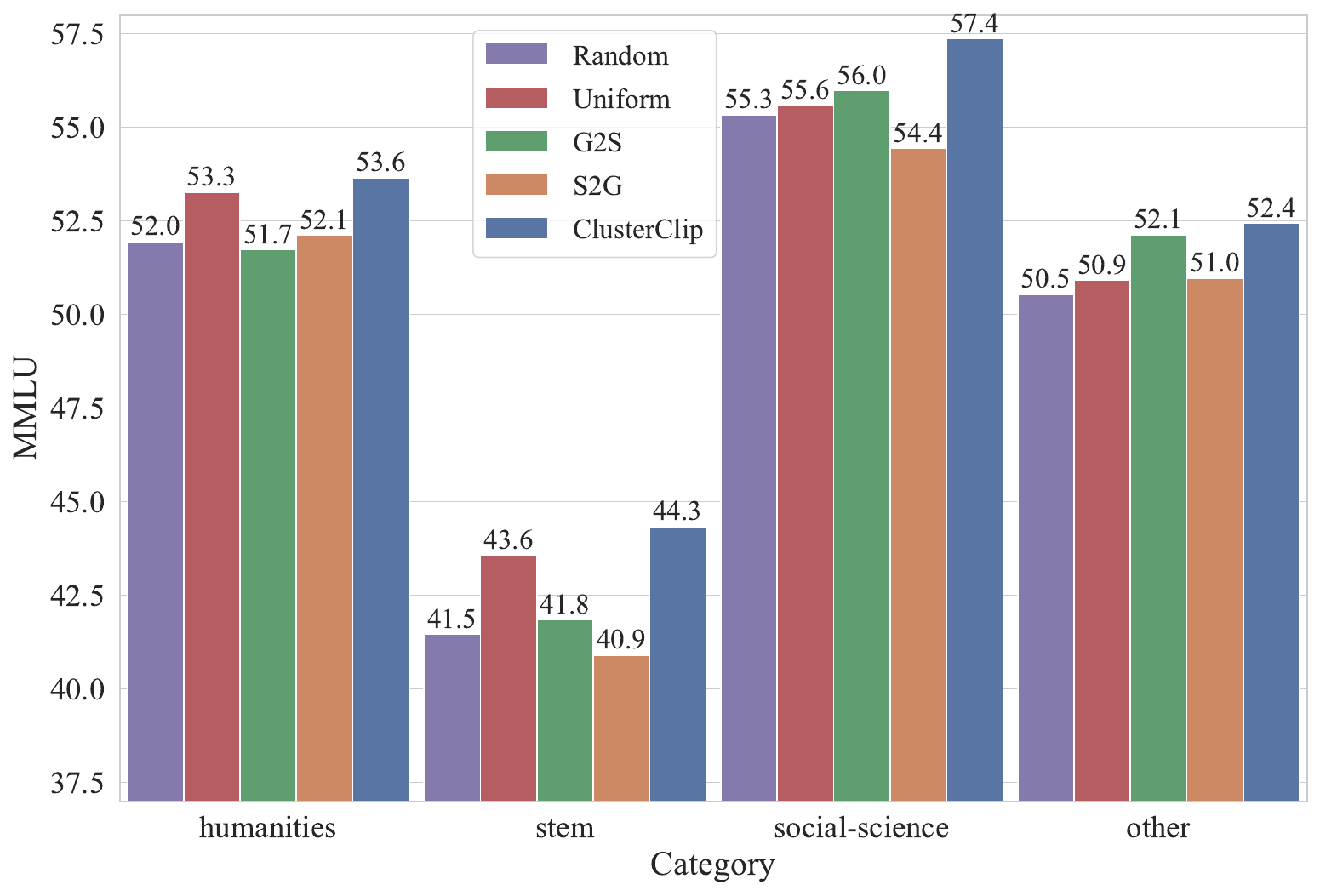}
    \caption{Accuracy of models trained with different sampling methods on each subset of MMLU.}
    \label{fig:result-proof-pile-mmlu}
\end{figure}

\paragraph{Results on Different Domains}
We find that ClusterClip can also boost the general performance of large language models on different domains. As shown in Figure~\ref{fig:result-proof-pile-mmlu}, the model trained with ClusterClip improves scores of all subsets of MMLU. The results on MMLU demonstrate that the model trained with ClusterClip can generalize across diverse tasks and domains even though the training set Proof-Pile-2 mainly targets mathematical tasks. Besides, Uniform Sampling also improves the performance when compared with Random Sampling, but still underperforms ClusterClip in all the categories of MMLU. Notably, the special cluster-based sampling methods (G2S and S2G Sampling) do not generalize well across different subsets of MMLU, which even underperform Random Sampling on humanities, stem, or social-science categories. 

\begin{figure}[htbp]
    \centering
    \includegraphics[width=0.9\linewidth]{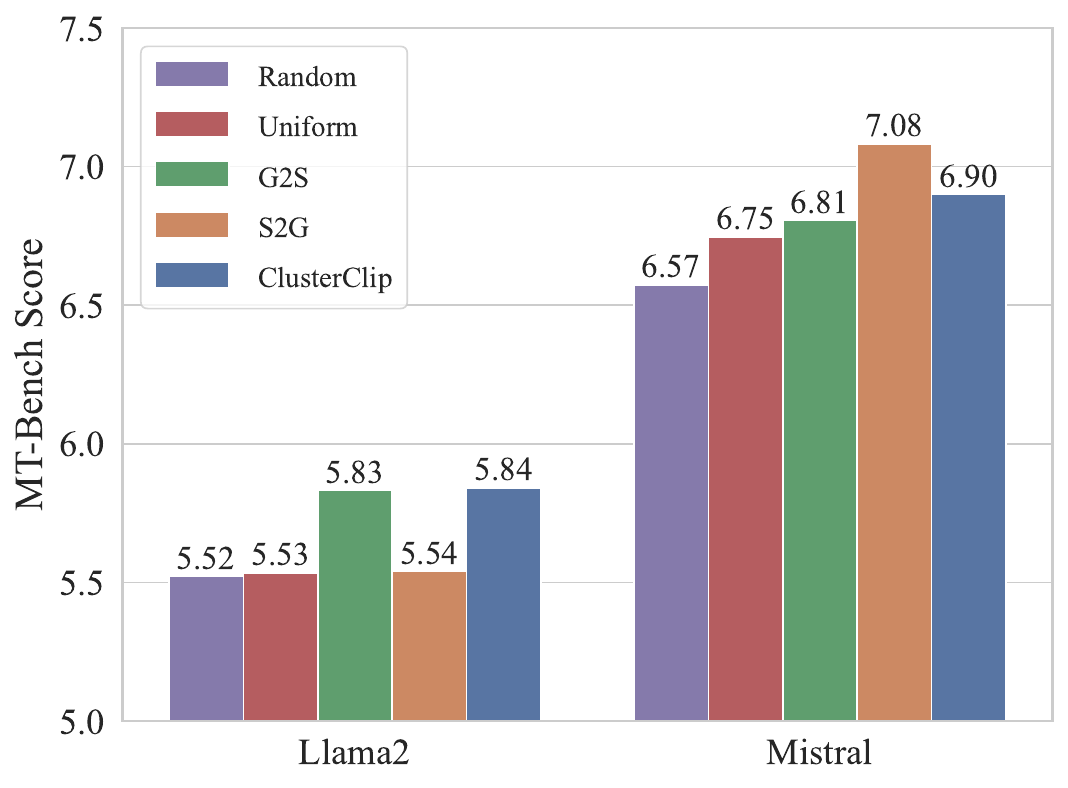}
    \caption{The results of different sampling methods on Llama2-7B and Mistral-7B, measured by MT-Bench Score.}
    \label{fig:result-diff-model-orca}
\end{figure}

\paragraph{Results on Different Models}
We also investigate the effectiveness of the proposed methods under different models. We train Llama2-7B on Open-Orca and compare the results with the performance of Mistral-7B. The training setups are kept the same in the supervised fine-tuning setups but the peak learning rate is set to 1e-5 to meet the requirements of fine-tuning Llama2-7B. As shown in Figure~\ref{fig:result-diff-model-orca}, ClusterClip consistently outperforms Random and Uniform Sampling by a large margin on both Llama2 and Mistral. However, the other three cluster-based sampling variants only improve marginal performance on different models, compared with Random Sampling.

\subsection{Analysis ClusterClip Sampling in Training}\label{sec:analysis-variants}
\begin{figure}[htbp]
    \centering
    \includegraphics[width=0.9\linewidth]{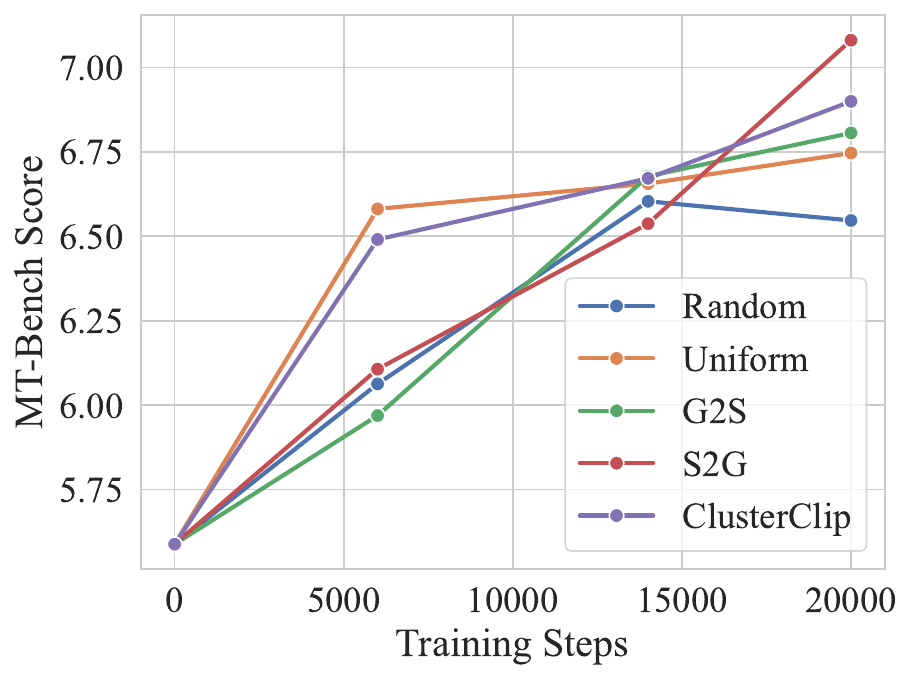}
    \caption{MT-Bench Scores of different sampling strategies during the training on Open-Orca dataset.}
    \label{fig:result-open-orca-mtbench}
\end{figure}

\begin{figure}[htbp]
    \centering
    \includegraphics[width=0.9\linewidth]{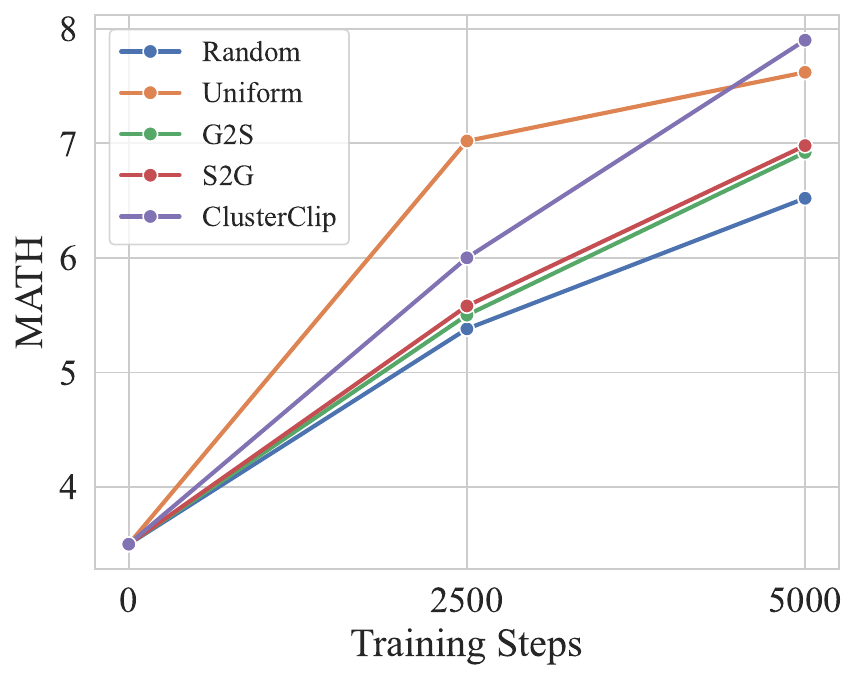}
    \caption{MATH Accuracy of different sampling strategies as the training progresses on the Proof-Pile-2 dataset.}
    \label{fig:result-proof-pile-math}
    
\end{figure}

\begin{figure}[htbp]
    \centering
    \includegraphics[width=0.8\linewidth]{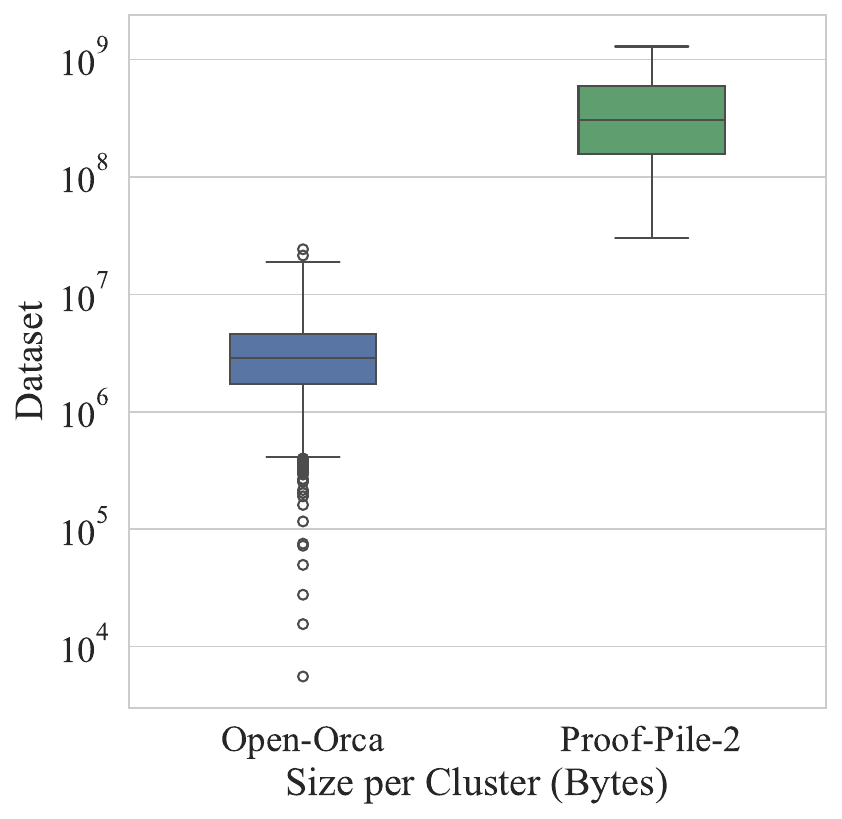}
    \caption{Distribution of cluster size (the number of bytes in each cluster) in different training sets. Clusters of Open-Orca have many outliers, especially tiny clusters, which could affect the sampling performance.}
    \label{fig:boxplot-dist}
    
\end{figure}

To find out how the ClusterClip Sampling affects the model performance as the training goes on, we massively evaluate the intermediate checkpoints of models trained with different sampling strategies. The supervised fine-tuning results are shown in Figure~\ref{fig:result-open-orca-mtbench} and the pre-training results are shown in Figure~\ref{fig:result-proof-pile-math}. We also present the data distribution of the training set in Figure~\ref{fig:boxplot-dist} to see the connection between the data distribution and the results of these sampling methods.

\paragraph{Analysis on Supervised Fine-Tuning}

As shown in Figure~\ref{fig:result-open-orca-mtbench}, all sampling strategies based on clustering outperform Random sampling, which demonstrates that cluster information can provide insights for sampling strategy design. Moreover, Random sampling improves the instruction-following ability at first but is quickly saturated and even overfitting at the end of the training. This indicates that Random sampling is unstable and sub-optimal for Open-Orca fine-tuning. 
Comparing clustering-based sampling methods, we surprisingly find that all these methods consistently improve the MT-Bench score across the training phases. Uniform sampling achieves good performance in the early stage but improves slowly, which may result from cluster repetition as the training for a large number of iterations. ClusterClip Sampling outperforms Uniform Sampling by a large margin at the end of the training, which indicates the effectiveness of the clip operation as the training goes on.
Surprisingly, both G2S and S2G outperform Uniform sampling at the end of training. S2G sampling even outperforms other sampling methods and shows a tendency to continue increasing the score for longer training. We assume that it benefits from the nature of Open-Orca, as the S2G sampling first learns large clusters followed by other clusters. These large clusters provide dense supervision in specific domains that helps the model quickly align to the instruction-following style in these domains and then transfer to more diverse domains and distributions. 

\paragraph{Analysis on Pre-Training}
However, the results of continual pre-training are different from the results of supervised fine-tuning. 
As shown in Figure~\ref{fig:result-proof-pile-math}, the random sampling under-performs compared with these cluster-based sampling strategies. Uniform sampling outperforms both G2S and S2G sampling methods, which is inconsistent with the results of supervised fine-tuning. 
It is the data distribution and the repetition rate of data samples in different datasets that lead to the divergence. 

\paragraph{Connection to Data Distribution}

\begin{figure}[htbp]
     \centering
     \begin{subfigure}[b]{0.23\textwidth}
         \centering
         \includegraphics[width=\textwidth]{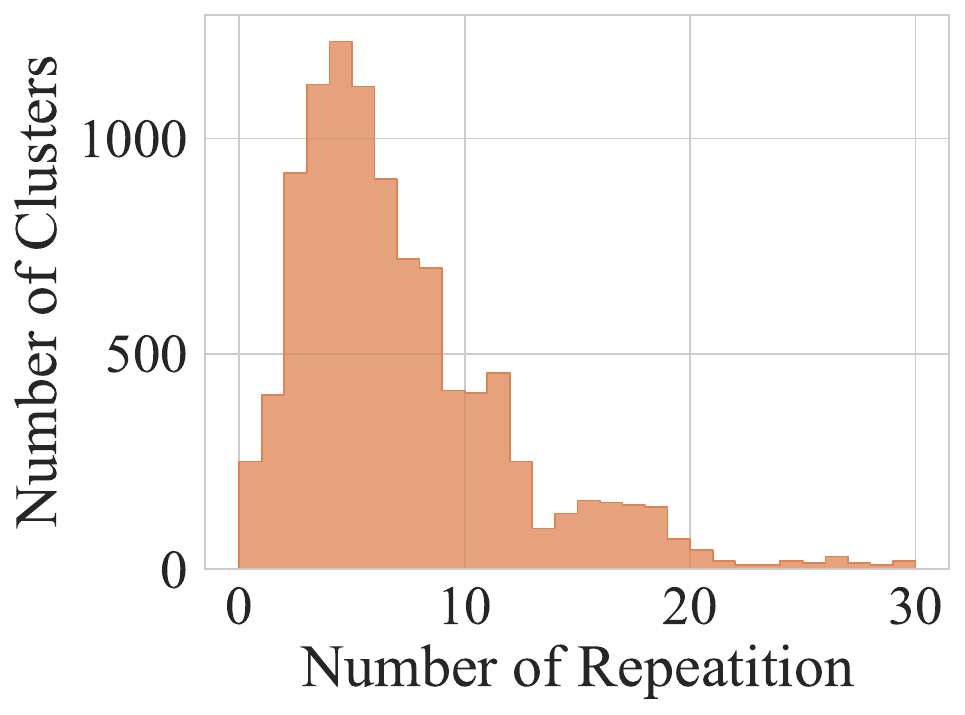}
         \caption{Open-Orca}
         \label{fig:open-orca-clusterepoch}
     \end{subfigure}
     \begin{subfigure}[b]{0.23\textwidth}
         \centering
         \includegraphics[width=\textwidth]{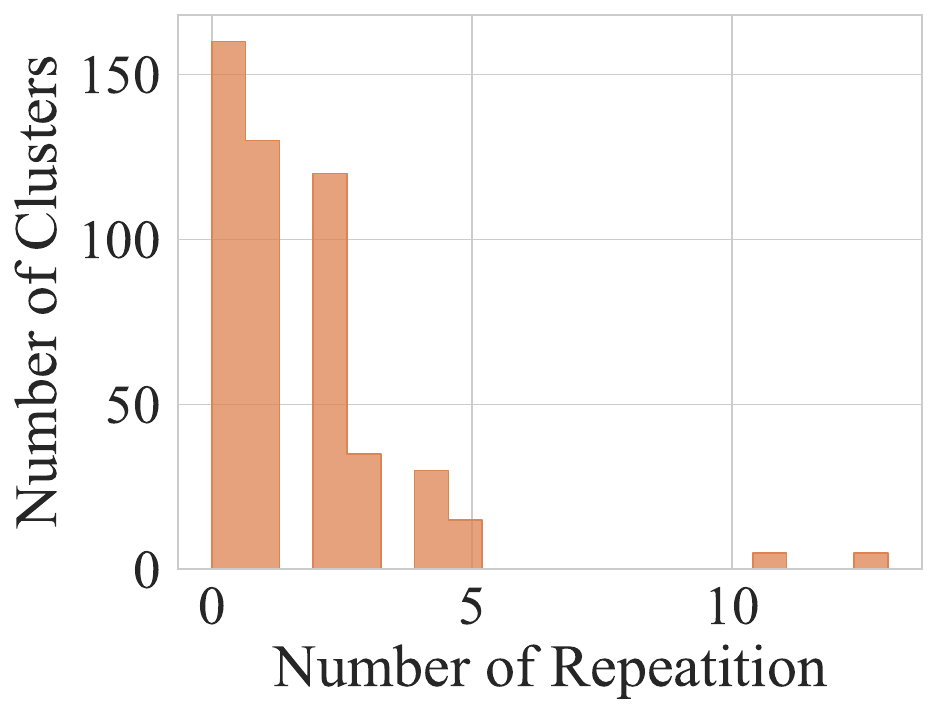}
         \caption{Proof-Pile-2}
         \label{fig:proof-pile-clusterepoch}
     \end{subfigure}
    \caption{Distribution of sample repetition on different clusters in Open-Orca and Proof-Pile-2 datasets.}
    \label{fig:training-clusterepoch}
        
\end{figure}

As the distribution of cluster sizes of Open-Orca and Proof-Pile-2 shown in Figure~\ref{fig:boxplot-dist}, both training sets have a bell shape that indicates a nearly normal distribution of document sizes, with a long tail of large-size clusters. While Open-Orca has a lot of clusters that have similar sizes, it contains some outlier clusters, like several huge clusters and many tiny clusters. We further calculate the repeated times of samples in these datasets when using Uniform Sampling and visualize the distribution of data repetition in Figure~\ref{fig:training-clusterepoch}. 
The repetition leads to the model over-fitting on these clusters and side effects on the overall performance of the model. The clusters in Open-Orca have been trained up to more than 30 epochs while the clusters in Proof-Pile-2 have been trained for at most 14 epochs. Thus the overfitting of Uniform Sampling is not severe on Proof-Pile-2, which makes the model generalize well on downstream mathematical tasks. However, the proposed ClusterClip Sampling still outperforms Uniform Sampling in Proof-Pile-2, indicating that the overfitting of rare documents still affects the model performance. The clip operation of ClusterClip effectively mitigates the overfitting of these texts, leading to performance gains on MATH and MMLU tasks.

It is worth noting that the training order of the samples may need to be carefully scheduled depending on different datasets, based on the results of G2S and S2G Sampling. It can be related to domain-specific learning or curriculum learning~\cite{bengio2009curriculum,hacohen2019curriculumDeep,roziere2023codeLlama}, which can be investigated in future work. And the cluster-based sampling methods are worth further exploration in this area.

\subsection{Ablation of ClusterClip Sampling}\label{sec:ablation}
We conduct experiments on Proof-Pile-2 to demonstrate how different configuration of ClusterClip Samping affects the overall performance.
\paragraph{Clip Threshold}

\begin{figure}[htbp]
    \centering
    \includegraphics[width=0.8\linewidth]{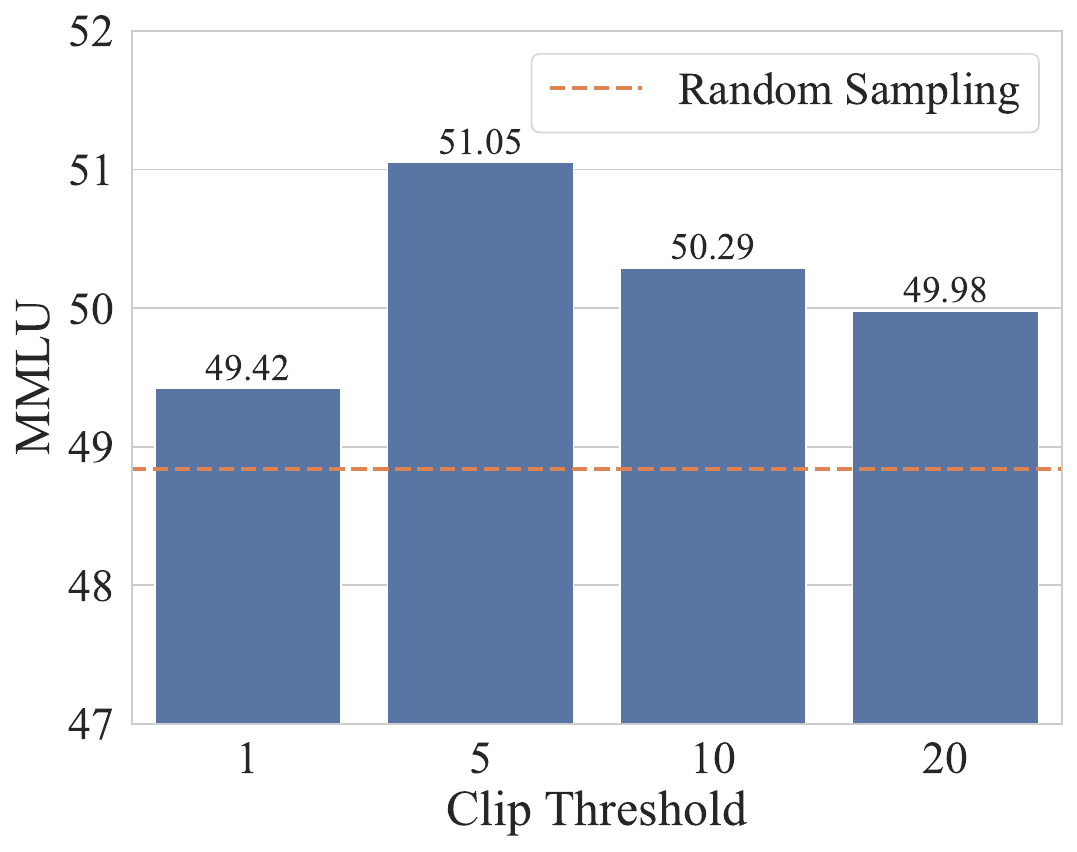}
    \caption{MMLU Accuracy of ClusterClip Sampling when changing the number of clipping thresholds on Proof-Pile-2.}
    \label{fig:ablation-cutoff-epoch}
\end{figure}

The number of maximum repetitions of one sample in ClusterClip Sampling is primarily manually set to 5, as suggested by~\citet{Muennighoff2023dataRepeat}. Nonetheless, we aim to verify that our setup is effective in reducing the overfitting on certain clusters. We train Llama2-7B on Proof-Pile-2 datasets with various clip thresholds of repeated samples and keep other setups the same. We provide the results in Figure~\ref{fig:ablation-cutoff-epoch}, where the too-small (only one time) or too-large (more than 10 times to repeat) clip thresholds degenerate the performance. Moreover, as the clip threshold increases, the MMLU performance of the trained LLM slightly decreases, which indicates that the overfitting issue becomes worse when samples are repeated more times. Thus, the value of 5 is near optimal for the threshold of the clipping.

\begin{figure}[htbp]
    \centering
    \includegraphics[width=0.8\linewidth]{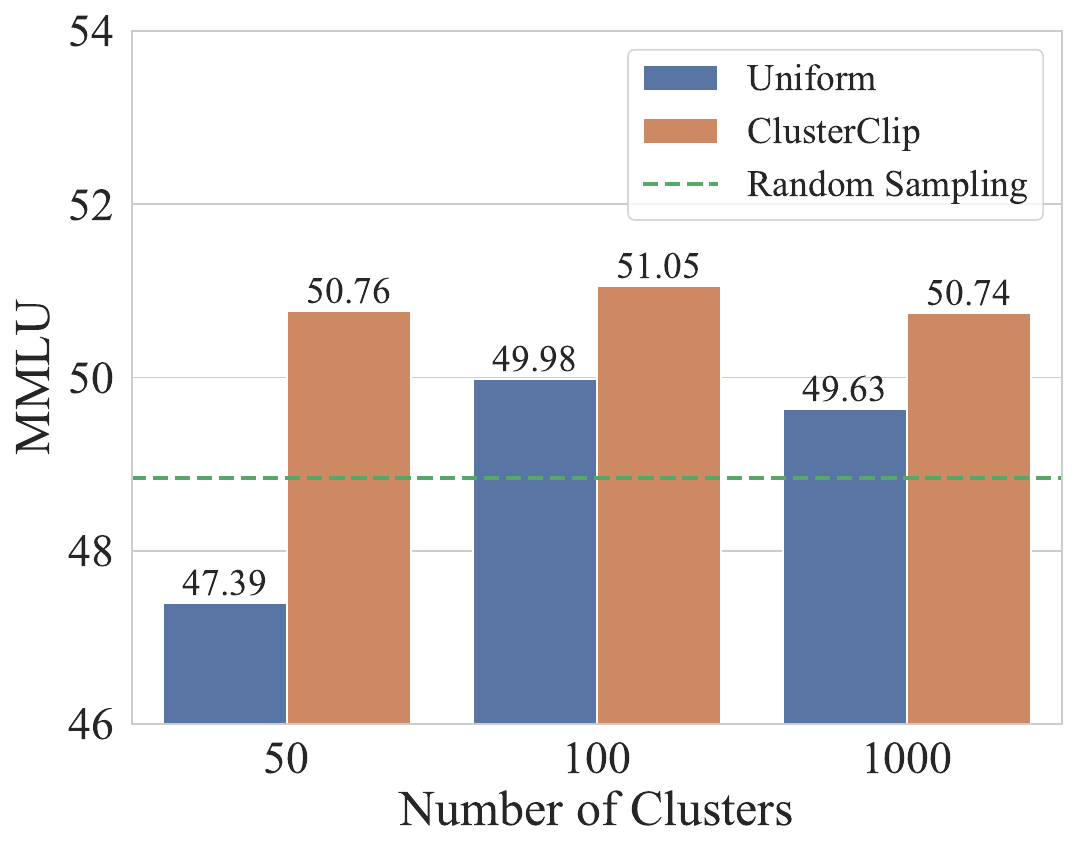}
    \caption{MMLU Accuracy of different sampling methods when changing the number of clusters on the Proof-Pile-2 dataset.}
    \label{fig:ablation-number-cluster}
\end{figure}

\paragraph{The Number of Clusters}
We aim to investigate the impact of the number of clusters in the application of ClusterClip Sampling methods. We conduct training of Llama2-7B on Proof-Pile-2 datasets with different numbers of clusters, following other setups in the main experiments. The results are shown in Figure~\ref{fig:ablation-number-cluster}, in which the ClusterClip sampling consistently outperforms Uniform sampling by a large margin on the MMLU benchmarks, even though the number of clusters is changing from 50 to 1000, respectively. Moreover, the number of clusters does not affect the performance much under the same cluster-based sampling methods. Thus, the ClusterClip sampling is not sensitive to the number of clusters.

\section{Conclusion}
In this work, we propose \textbf{ClusterClip Sampling} based on data clustering to balance the long-tail distribution of the training set of large language models. We compare the proposed method with Random Sampling and other cluster-based sampling method variants in both supervised fine-tuning and pre-training. Extensive experimental results across 7 datasets across diverse tasks and domains demonstrate the effectiveness of ClusterClip Sampling, which outperforms baselines under different training sets and models. 
We hope this work can instigate more research on data sampling approaches for improving language model training.

\section*{Limitations}
The ClusterClip Sampling proposed in this work can improve the training of LLMs and mitigate overfitting on small clusters. However, there are limitations in our current work and we hope to
enhance the framework of data sampling in future research. 
Firstly, We use transformer-based sentence embedding to generate data representation and exploit K-Means for clustering. Nonetheless, other data representation or clustering methods can be incorporated. For example, using an LLM to process the texts to achieve better clustering accuracy.
Secondly, our current method samples data mainly based on cluster size. However, extra model information or dataset statistics can be incorporated for better sampling strategies. Future work should explore more sophisticated methods to determine the document-level or token-level sampling probabilities.
Finally, this study concentrates on language models that only process texts. Training multi-modal generative models that can understand and generate images, videos, and audio poses its challenges. And it requires more sophisticated data processing and sampling techniques, which can be explored in the future.

\section*{Ethics Statement}
We leveraged data clustering to find semantic clusters of training data and utilize the cluster information for data sampling. There is a risk that the cluster-based sampling aggregates the bias or toxic content in the dataset. However, in this work, we utilize open-sourced datasets to train the LLMs, including Open-Orca~\cite{lian2023openOrca} and Proof-Pile-2~\cite{azerbayev2023llemma}. These datasets are specifically collected and filtered to avoid toxic and safety issues. When applying the cluster-based sampling strategies for new datasets, data cleaning, and filtering methods can be used to remove the harmful content before clustering and training.

\section*{Acknowledgements}
This work was supported by the National Key Research and Development Program of China (No.2022ZD0160102).  The computations in this research were performed using the CFFF platform of Fudan University.
\bibliography{anthology,custom}
\bibliographystyle{acl_natbib}




\end{document}